\documentclass[letterpaper]{article} % DO NOT CHANGE THIS
\usepackage[preprint]{aaai2027}  % DO NOT CHANGE THIS
\usepackage[hyphens]{url}  % DO NOT CHANGE THIS
\usepackage{graphicx} % DO NOT CHANGE THIS
\usepackage{natbib}  % DO NOT CHANGE THIS AND DO NOT ADD ANY OPTIONS TO IT
\usepackage{caption} % DO NOT CHANGE THIS AND DO NOT ADD ANY OPTIONS TO IT
\usepackage{algorithm}
\usepackage{algorithmic}
\usepackage{multirow}
\usepackage{tabularx}
\usepackage{amsmath}
\usepackage{newfloat}
\usepackage{listings}
\DeclareCaptionStyle{ruled}{labelfont=normalfont,labelsep=colon,strut=off} % DO NOT CHANGE THIS
\floatstyle{ruled}
\newfloat{listing}{tb}{lst}{}
\floatname{listing}{Listing}

\usepackage{booktabs}
\usepackage[most]{tcolorbox}

\newtcolorbox{promptbox}[1]{
  enhanced,
  breakable,
  colback=gray!5,
  colframe=black!55,
  boxrule=0.5pt,
  arc=0pt,
  top=2mm, bottom=2mm, left=2mm, right=2mm,
  fonttitle=\bfseries\small,
  fontupper=\ttfamily\footnotesize,
  attach boxed title to top left={yshift=-2mm, xshift=2mm},
  boxed title style={
    enhanced, arc=2mm, boxrule=0pt,
    colback=black!70, colframe=black!70,
    top=0.8mm, bottom=0.8mm, left=2mm, right=2mm,
  },
  coltitle=white,
  title={#1},
  top=6mm,
}

\title{Beyond a Single Judge: The Evidence-Grounded, Social-Weighted\\ Persona Panel for Generative UI Evaluation}
\author{
    Zheng Wu\textsuperscript{\rm 1},
    Yibo Luo\textsuperscript{\rm 1},
    Pu Zhang\textsuperscript{\rm 1},
    Cheng Yang\textsuperscript{\rm 2},
    Zhuosheng Zhang\textsuperscript{\rm 1}\corresponding
}
\affiliations{
    \textsuperscript{\rm 1}School of Computer Science, Shanghai Jiao Tong University\\
    \textsuperscript{\rm 2}ByteDance Inc\\
    wzh815918208@sjtu.edu.cn, zhangzs@sjtu.edu.cn
}

\begin{document}

\maketitle

\begin{abstract}
Generative UI (GenUI) lets large language models synthesize a complete, renderable interface directly from a natural-language instruction, but evaluating the quality of what they generate remains an open problem. 
Human evaluation is costly and rater-variant, while LLM-as-a-judge is scalable but reflects only a single implicit viewpoint, unable to capture how different populations of real users actually perceive the same interface. 
We propose the Evidence-Grounded, Social-Weighted Persona Panel (ESPP), a three-stage GenUI evaluation method in which a panel of psychologically diverse, evidence-grounded personas independently rates a screenshot, exchanges opinions under a trait-derived, semantically-gated bounded-confidence mechanism, and is aggregated via Delphi-inspired social weighting into a single judgment. 
ESPP tracks human judgment substantially more closely than a naive single-pass judge, raising Pearson $r$ from $0.716$ to $0.922$, and a prompt-ensemble control recovers only about a third of this gap, isolating genuine persona and evidence grounding as the dominant source of improvement.
Beyond this fidelity gain, retaining each panelist's individual rating further reveals that user subgroups agree on overall model rankings yet diverge sharply on specific rating dimensions, a structural disagreement a single homogeneous judge would systematically erase.
The codes are available at \url{https://github.com/Wuzheng02/ESPP}.
\end{abstract}

\section{Introduction}

As large language models (LLMs) grow increasingly capable of multi-step reasoning and tool use~\cite{yao2022react}, they are no longer confined to producing text: given a natural-language instruction, an LLM can now directly generate a complete, renderable user interface tailored to that request~\cite{chen2026generative,leviathan2026generative,wang2025generative}. 
The generative UI (GenUI) paradigm, promises to collapse the traditional design-to-implementation pipeline into a single generative step, letting an interface be synthesized on demand for a specific user's need~\cite{lieberman2006end}.

\begin{figure}[t]
\centering
\includegraphics[width=1\linewidth]{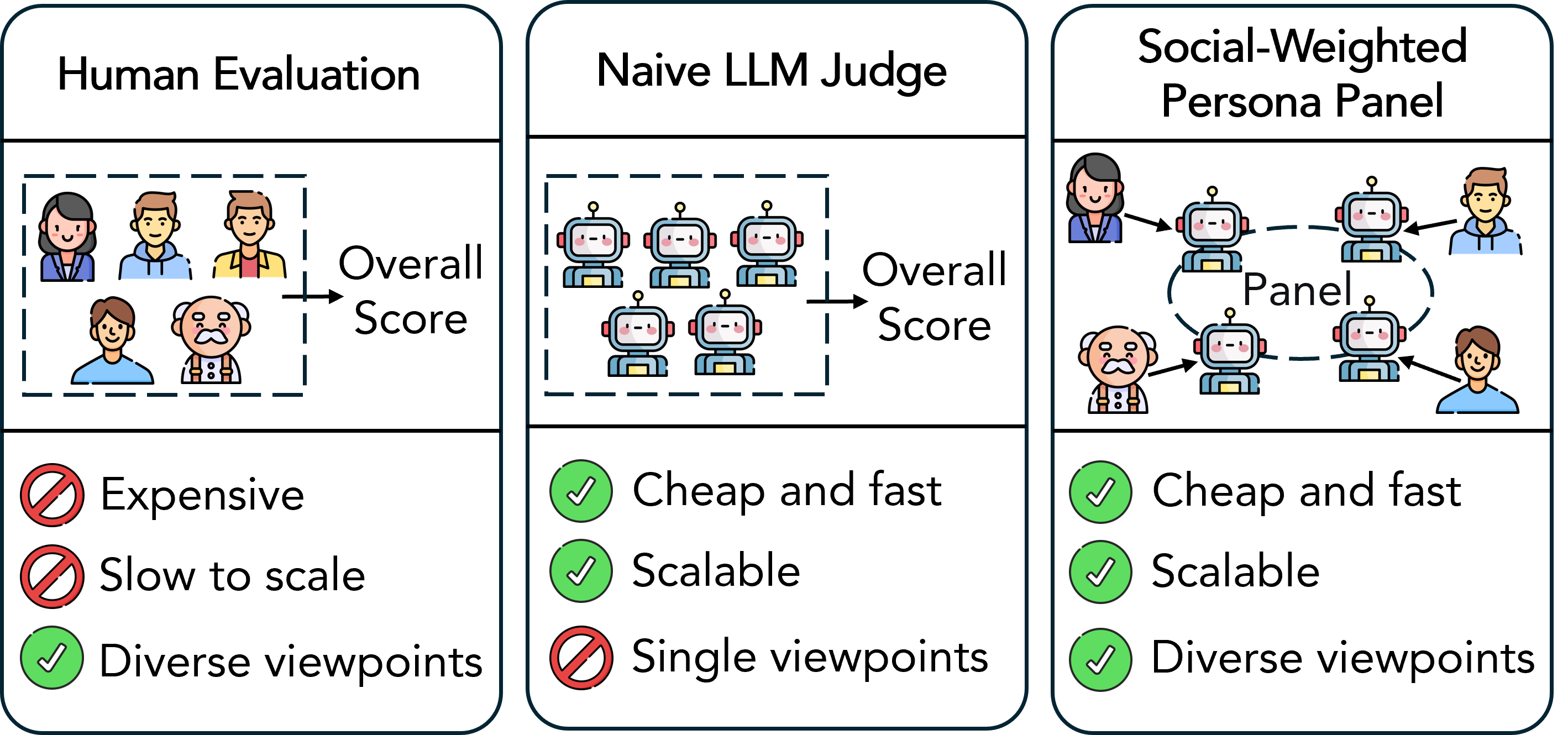}
\caption{Three paradigms for evaluating Generative UI. Human evaluation is diverse but costly; a naive LLM judge is scalable but reflects only a single viewpoint and tracks human judgment poorly ($r=0.72$); our Evidence-Grounded, Social-Weighted Persona Panel combines both, tracking human judgment far more closely ($r=0.92$).}
\label{fig:teaser}
\end{figure}

However, evaluating the quality of a generated UI remains an open problem. As shown in Figure~\ref{fig:teaser}, existing GenUI evaluation falls into two broad categories: human evaluation and LLM-as-a-judge. Human evaluation is costly and slow to scale, and its outcomes carry substantial rater-to-rater variance~\cite{clark2021all}. LLM-as-a-judge is fast and cheap~\cite{zheng2023judging}, but a single judging pass reflects only one implicit viewpoint, lacking the diversity needed to represent how different populations of real users actually perceive the same interface; even efforts to diversify the judge by pooling several models or samples still sample repeatedly from LLM-shaped viewpoints, falling short of a population of genuinely distinct users~\cite{verga2024replacing}.

To address this gap, we propose the \textbf{Evidence-Grounded, Social-Weighted Persona Panel} (ESPP), a three-stage GenUI evaluation method in which a panel of psychologically diverse~\cite{serapio2023personality}, evidence-grounded personas independently rates a screenshot, exchanges opinions under a semantic bounded-confidence mechanism~\cite{hegselmann2002opinion}, and is aggregated via Delphi-inspired social weighting~\cite{dalkey1963delphi,bail2024can} into a single judgment.

We validate ESPP on UIPersonaBench, a benchmark we construct of 500 UI-generation instructions rendered by 14 state-of-the-art models against human ground truth. 
ESPP tracks human judgment substantially more closely than a naive single-pass LLM judge, raising Pearson $r$ from $0.716$ to $0.922$, and this gain cannot be fully explained by an artifact of averaging multiple LLM calls alone: a prompt-ensemble control that averages 5 independently-prompted passes recovers only about a third of this gap, showing that most of the improvement comes from genuine persona and evidence grounding, with sampling variance playing only a minor role. 
Beyond this headline result, our most consequential further finding is that different user subgroups agree closely on an overall leaderboard yet diverge sharply on specific dimensions such as Control~\cite{sunstein2002law}: a single homogeneous judge would report only the former and silently erase the latter, whereas our panel retains and reports both. We additionally confirm that the panel's psychological mechanisms are not decorative: trait-derived receptivity measurably predicts how much a persona revises its rating during deliberation, panel discussion converges only partially instead of collapsing to consensus, and social-weighted aggregation is no more susceptible to cosmetic dark-pattern manipulation than a naive judge.

In summary, we make the following contributions:

(i) We propose the Evidence-Grounded, Social-Weighted Persona Panel, a three-stage GenUI evaluation method that grounds each persona's rating in its own prior evidence, models opinion exchange via a trait-derived, semantically-gated bounded-confidence mechanism, and aggregates panelist judgments with Delphi-inspired social weighting.

(ii) We construct UIPersonaBench, a 500-instruction, 14-model, 7,000-screenshot GenUI benchmark with paired human ground truth, enabling systematic comparison of automatic evaluation methods against real human judgment.

(iii) We show that ESPP improves fidelity to human judgment over a naive single judge and a prompt-ensemble control, isolating genuine persona evidence grounding as the dominant source of this improvement.

(iv) Through a series of further analyses, we surface real, dimension-localized disagreement across user subgroups that a single homogeneous judge would otherwise obscure, and provide behavioral evidence that our panel's psychological mechanisms operate as designed rather than as decorative prompt instructions.

\section{Related Work}

In this section, we review prior work on generative UI and its evaluation, and on LLM-as-a-judge methods for automatic quality assessment, situating our Evidence-Grounded, Social-Weighted Persona Panel relative to both lines of work.

\subsection{Generative UI}
Generative UI extends LLMs from producing plain text to directly synthesizing renderable interface artifacts~\cite{chen2026generative,leviathan2026generative,wang2025generative}. 
This spans front-end code generation from natural-language or sketch specifications~\cite{yang2025ui2code}, layout and component synthesis conditioned on design constraints~\cite{feng2023layoutgpt}, and end-to-end web/app generation pipelines that couple an LLM with a rendering or compilation backend~\cite{si2025design2code}. 
As these systems mature from toy demos to models capable of one-shot, production-quality UI synthesis, evaluating what they produce has become the bottleneck: most GenUI work still reports functional correctness or code-level metrics, or relies on a handful of human raters on a small artifact set~\cite{wu2024uiclip}, leaving open how to assess subjective, user-facing qualities such as trust, control, and transparency at scale and across a population of users with heterogeneous needs.

\subsection{LLM-as-a-Judge}
Using an LLM to score another model's output has become a standard substitute for human evaluation. 
Subsequent work has refined the judge itself via structured rubrics~\cite{kim2024prometheus}, critique generation~\cite{mcaleese2024llm}, and bias calibration~\cite{wang2024large,saito2023verbosity}, or reduced single-pass variance by aggregating multiple judges or samples~\cite{verga2024replacing,zheng2023judging}. 
A related thread instantiates LLMs as personas or synthetic agent populations to simulate social behavior~\cite{park2023generative}, survey responses~\cite{argyle2023out}, or multi-agent debate~\cite{du2024improving}, typically from a demographic or trait description alone~\cite{park2022social}. 
Our work departs from both threads. 
Prior work treats judge diversity as repeated sampling from one implicit viewpoint and treats a persona as an ungrounded trait label; we instead ground each persona's rating in its own documented evidence and structure inter-persona interaction with an explicit, psychologically-motivated opinion-dynamics mechanism, turning the judge population itself into a socially-weighted, behaviorally-grounded model of real user disagreement.

\section{Evidence-Grounded, Social-Weighted Persona Panel for Generative UI Evaluation}
\label{sec:method}

\begin{figure*}[t]
\centering
\includegraphics[width=1\linewidth]{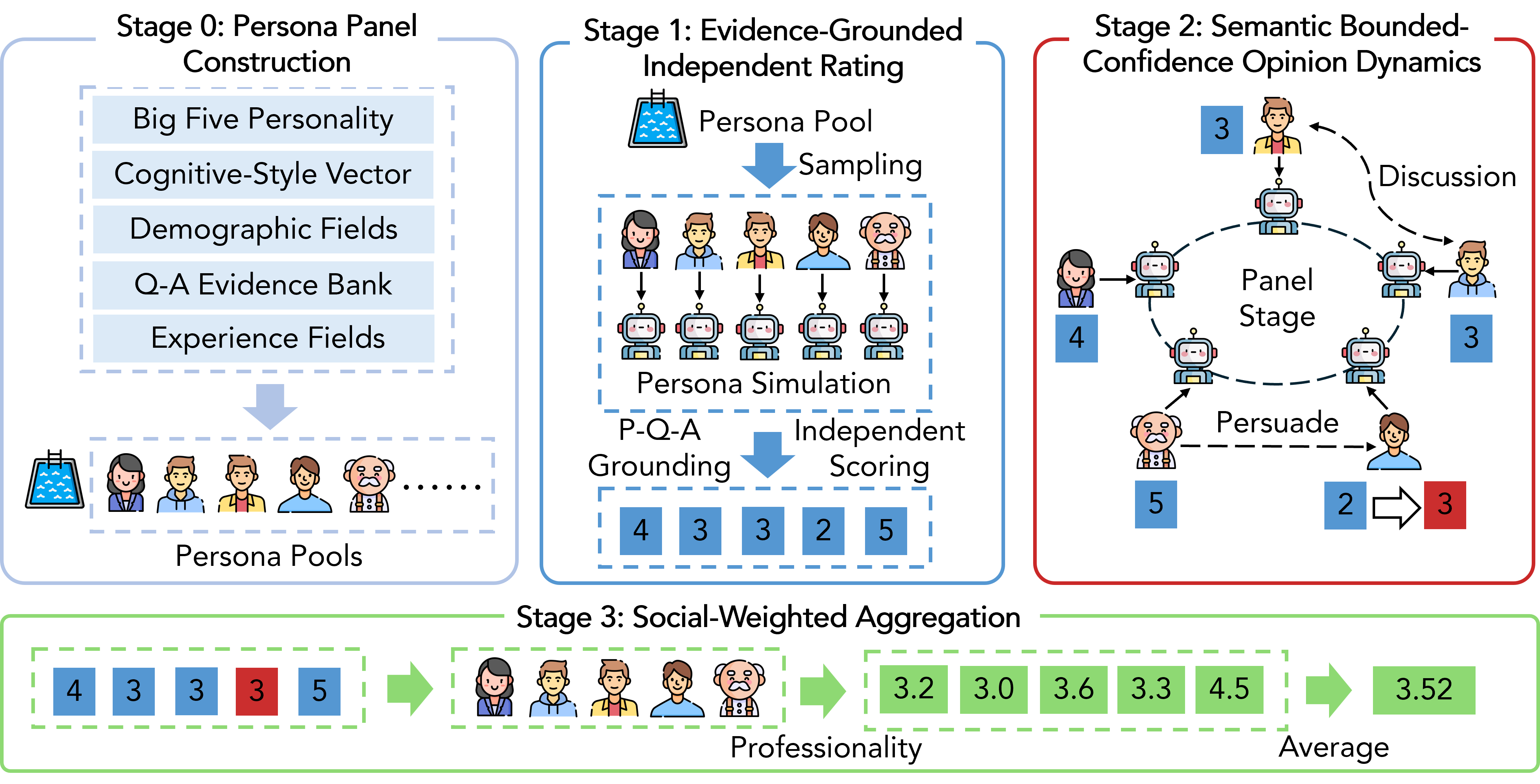}
\caption{Overview of the Evidence-Grounded, Social-Weighted Persona Panel. A diversity-constrained panel $\Pi_x$ is sampled from a pool of 1000 personas (Persona Panel Construction); each panelist independently rates screenshot conditioned on its P-Q-A evidence (Stage 1); panelists revise their ratings through semantic bounded-confidence opinion exchange modulated by trait-derived receptivity (Stage 2); and revised ratings are combined via weighted aggregation into the final score (Stage 3).}
\label{fig:pipeline}
\end{figure*}

In this section, as shown in Figure~\ref{fig:pipeline}, we introduce the Evidence-Grounded, Social-Weighted Persona Panel:
a Persona Panel Construction procedure followed by three stages that ground, discuss, and aggregate the judgments of a population of synthetic raters, namely independent evidence-grounded rating, semantic bounded-confidence opinion exchange, and social-weighted aggregation.

\subsection{Stage 0: Persona Panel Construction}
A single judge, however capable, is a sample of size one drawn from an unknown and almost certainly non-representative distribution over real users; treating repeated LLM calls as a substitute for population diversity conflates variance reduction with coverage. We instead parameterize a rater population along axes with independent construct validity in the personality and judgment-and-decision-making literatures. Each persona $p=(\mathbf{b},\mathbf{c},\mathbf{d},\mathbf{e},q)$ is a synthetic rater instantiated from a Big-Five trait vector $\mathbf{b}\in\{\mathrm{low,med,high}\}^5$ over openness, conscientiousness, extraversion, agreeableness, and neuroticism, grounded in the Five-Factor Model~\cite{costa2014neo}; a cognitive-style vector $\mathbf{c}$ specifying analytical-versus-intuitive information processing~\cite{epstein1996cognitive}, risk tolerance, and need for control; and demographic and experience fields $\mathbf{d},\mathbf{e}$ (age, occupation, tech literacy, domain expertise). To avoid leaving these fields as inert labels, each persona additionally carries a bank $q$ of persona-specific question-answer pairs $q=\{(\text{question}_k,\text{answer}_k)\}$, elicited once per persona per evaluation dimension, so that $(\mathbf{b},\mathbf{c},\mathbf{d},\mathbf{e})$ are behaviorally instantiated, not merely declared: a persona's disposition is evidenced by what it has already said, and an adjective alone is never enough.

A panel that merely samples $N$ personas uniformly at random risks two failure modes: near-duplicate viewpoints that waste panel capacity, or an all-idiosyncratic panel with no shared ground for the deliberation in Stage 2 to act on. We therefore draw, for each instruction $x$, a fixed panel $\Pi_x=\{p_1,\dots,p_N\}$ ($N=5$) via stratified sampling over $(\mathbf{d},\mathbf{b})$ from the 1{,}000-persona pool, subject to a diversity constraint on the trait-space distance $\delta(p_i,p_j)=\tfrac{1}{|\mathbf{b}|+|\mathbf{c}|}\lVert(\mathbf{b}_i,\mathbf{c}_i)-(\mathbf{b}_j,\mathbf{c}_j)\rVert_1$: $\Pi_x$ must contain at least one close pair (small $\delta$, a natural coalition) and one distant pair (large $\delta$, a latent fault line), so that Stage 2 is guaranteed both a consensus to reinforce and a disagreement to resolve, without degenerating into either unanimous agreement or unstructured noise. The same $\Pi_x$ is reused across every model rendering $x$, holding the jury fixed so that cross-model comparisons are never confounded by a change in who is judging.

\subsection{Stage 1: Evidence-Grounded Independent Rating}
Each $p_i\in\Pi_x$ independently rates screenshot $y$ on 5 GenUI dimensions $\mathcal{K}$ (Understanding, Trust\_and\_Reliance, Usability, Control, Transparency), producing $s^{(1)}_i\in\{1,\dots,5\}^{\mathcal{K}}$ with a natural-language justification per dimension. An ungrounded persona is free to rationalize any score: given only a trait label such as ``low agreeableness,'' an LLM can construct a post-hoc justification for essentially any rating, since the label constrains the conclusion far less than it constrains the surface style of the reasoning that leads there. To avoid eliciting $s^{(1)}_i$ from trait labels alone, we therefore close this degree of freedom at its source: the prompt for dimension $k$ is conditioned on $p_i$'s evidence $q_i^{(k)}\subset q_i$ elicited in Stage 0, and $p_i$ is instructed to keep $s^{(1)}_{i,k}$ consistent with $q_i^{(k)}$. 
This P-Q-A grounding anchors each rating in a documented, persona-specific behavioral history and steers it away from a context-free vibe; 
it is what lets Stage 2 attribute a subsequent rating change specifically to peer argument, and not to the persona simply re-rolling a loosely-constrained justification.

\subsection{Stage 2: Semantic Bounded-Confidence Opinion Dynamics}
Independent ratings alone cannot express deliberation: real panels revise views under argument, but not unconditionally, and a mechanism that lets every peer's opinion move every other peer collapses to a plain mean, while a mechanism that lets no opinion move anyone reduces Stage 2 to Stage 1 with extra steps. What is needed, instead, is a rule for exactly which arguments a given panelist should find persuasive. Classical bounded-confidence opinion dynamics~\cite{hegselmann2002opinion} supplies exactly this rule for scalar opinions, in which agent $i$ only integrates peer $j$'s opinion if $|s_i-s_j|<\epsilon$: nearby opinions merge, distant ones are ignored outright. Applied unmodified to LLM-backed personas, however, this scalar gate is both too permissive and too blind: it licenses influence from a numerically close but substantively irrelevant peer comment, and it cannot express that receptivity itself is a trait-dependent quantity, not a shared constant $\epsilon$. We therefore adapt the classical model to LLM-backed personas along two axes absent from the original formulation.

\textbf{Trait-derived receptivity.} Susceptibility to social influence is treated as a deterministic function of $p_i$'s own traits, not as a free parameter,
\begin{equation}
  \rho(p_i) = \rho_{\min} + \frac{a_i+n_i}{4}\bigl(\rho_{\max}-\rho_{\min}\bigr), \quad a_i,n_i\in\{0,1,2\},
  \label{eq:receptivity}
\end{equation}
where $a_i,n_i$ are $p_i$'s ordinal agreeableness and neuroticism levels: higher agreeableness (desire for group harmony) and higher neuroticism (sensitivity to social pressure) both raise $\rho(p_i)$, consistent with FFM correlates of persuadability~\cite{costa2014neo}. $\rho(p_i)$ is injected as a qualitative behavioral instruction, not a numeric weight the LLM computes with.

\textbf{Semantic gate.} Beyond the numeric gate $|s_i-s_j|<\epsilon$, a peer's argument is only licensed to move $p_i$ if it is on-topic for $p_i$: it must reference a concern in $p_i$'s own topical hook set $H_i$ (their stated frustration, UI expectations, and control/risk disposition). An on-topic argument can shift $p_i$ across a wide score gap, while an off-topic one does not move $p_i$ even from a narrow gap, capturing an assimilation-contrast asymmetry from social judgment theory~\cite{sherif1961social} that a purely numeric gate cannot express. Each $p_i$ observes all peer ratings and reasoning for every $k\in\mathcal{K}$ simultaneously and outputs a revised $s^{(2)}_i$.

\subsection{Stage 3: Social-Weighted Aggregation}
A plain mean of $\{s^{(2)}_i\}$ treats every panelist as equally informative, ignoring that some personas are more expert or more representative of an instruction's target use case; yet the opposite extreme, deferring entirely to a single self-declared expert, would simply reintroduce the single-viewpoint failure mode of a naive judge one level up, this time at the aggregation stage instead of the rating stage. The Delphi method addresses this exact tension in human expert elicitation by iteratively reweighting toward more informative panelists while still pooling the full panel~\cite{dalkey1963delphi}; we adapt its spirit to a single-round LLM setting via a soft, closed-form reweighting, leaving aside its iterative questionnaire mechanics: panelist $p_i$'s weight for instruction $x$ (scenario $c$) is
\begin{equation}
  w_i \;\propto\; 1 + \lambda_e\, e(p_i) + \lambda_r\, r(p_i, c),
  \label{eq:weight}
\end{equation}
normalized to $\sum_i w_i=1$ over $\Pi_x$, where $e(p_i)\in[0,1]$ is an expertise score averaging $p_i$'s domain experience and tech literacy, and $r(p_i,c)\in[0,1]$ is a representativeness score combining $p_i$'s target-domain match to scenario $c$ with their usage frequency. $\lambda_e=\lambda_r=0.5$ keep the reweighting mild by construction, so no single persona dominates the aggregate: $w_i$ can shift the panel's emphasis but, unlike a hard veto or a top-1 selection rule, cannot silence a dissenting panelist outright. The final per-dimension score is $\hat{s}_k=\sum_i w_i\, s^{(2)}_{i,k}$, and the overall score is the mean of $\hat{s}_k$ over $\mathcal{K}$. Crucially, aggregation collapses $\Pi_x$'s judgments into a single number for benchmarking purposes, but it does not discard the disaggregated per-persona ratings $\{s^{(2)}_i\}$ that produced it.
% Section~\ref{sec:divergence} shows why retaining them matters.

\begin{table}[t]
\centering
\begin{tabular}{lcccc}
\toprule
Evaluation method & $r$ & $\rho$ & MAE $\downarrow$ & RMSE $\downarrow$ \\
\midrule
Naive single-pass & 0.716 & 0.612 & 0.435 & 0.490 \\
Prompt-ensemble& 0.789 & 0.715 & 0.277 & 0.329 \\
ESPP& \textbf{0.922} & \textbf{0.904} & \textbf{0.110} & \textbf{0.138} \\
\bottomrule
\end{tabular}
\caption{Alignment between each evaluation method and human ground truth at the individual-screenshot level.}
\label{tab:judge_fidelity}
\end{table}
\begin{table*}[t!]
\centering
\setlength{\tabcolsep}{3pt}
\begin{tabular}{lcccccccccccc}
\toprule
\multirow{2}{*}{Model} & \multirow{2}{*}{Overall} & \multicolumn{5}{c}{Dimension} & \multicolumn{6}{c}{Scenario} \\
\cmidrule(lr){3-7} \cmidrule(lr){8-13}
 & & Underst. & Trust & Usab. & Ctrl & Transp. & Landing & Dashboard & Forms & Ecomm. & Social & Mobile \\
\midrule
Claude-Opus-4.7 & \textbf{3.73} & 4.83 & \textbf{3.61} & 3.94 & 3.45 & 2.82 & 3.65 & \textbf{3.67} & \textbf{3.88} & 3.72 & 3.70 & \textbf{3.77} \\
GPT-5.4 & 3.72 & 4.85 & 3.60 & 3.96 & 3.43 & 2.78 & \textbf{3.67} & 3.65 & 3.86 & \textbf{3.74} & 3.69 & 3.74 \\
GPT-5.5 & 3.72 & \textbf{4.86} & \textbf{3.61} & \textbf{3.97} & 3.40 & 2.77 & 3.66 & 3.67 & 3.86 & 3.73 & \textbf{3.70} & 3.72 \\
GPT-5.2 & 3.67 & 4.66 & 3.54 & 3.77 & \textbf{3.48} & \textbf{2.89} & 3.57 & 3.67 & 3.80 & 3.64 & 3.61 & 3.70 \\
DeepSeek-V4-Pro & 3.66 & 4.75 & 3.53 & 3.89 & 3.41 & 2.75 & 3.65 & 3.63 & 3.77 & 3.65 & 3.62 & 3.67 \\
DeepSeek-V4-Flash & 3.64 & 4.72 & 3.50 & 3.90 & 3.40 & 2.71 & 3.58 & 3.62 & 3.80 & 3.55 & 3.63 & 3.68 \\
Claude-Opus-4.6 & 3.64 & 4.75 & 3.50 & 3.88 & 3.38 & 2.69 & 3.57 & 3.62 & 3.77 & 3.60 & 3.60 & 3.69 \\
GLM-5.1 & 3.63 & 4.71 & 3.51 & 3.82 & 3.38 & 2.71 & 3.57 & 3.59 & 3.74 & 3.61 & 3.58 & 3.67 \\
Doubao-Seed-2.0-Pro & 3.62 & 4.75 & 3.42 & 3.90 & 3.38 & 2.64 & 3.52 & 3.61 & 3.74 & 3.58 & 3.60 & 3.66 \\
GLM-5 & 3.62 & 4.72 & 3.49 & 3.86 & 3.38 & 2.64 & 3.52 & 3.61 & 3.70 & 3.56 & 3.65 & 3.67 \\
MiniMax-M2.7 & 3.56 & 4.65 & 3.41 & 3.78 & 3.34 & 2.59 & 3.51 & 3.53 & 3.66 & 3.54 & 3.51 & 3.58 \\
GLM-4.7 & 3.50 & 4.60 & 3.32 & 3.81 & 3.30 & 2.48 & 3.50 & 3.49 & 3.61 & 3.47 & 3.43 & 3.51 \\
Kimi-K2.6 & 3.48 & 4.51 & 3.37 & 3.65 & 3.31 & 2.58 & 3.29 & 3.50 & 3.62 & 3.50 & 3.49 & 3.50 \\
Gemini-2.5-Pro & 3.45 & 4.60 & 3.26 & 3.77 & 3.24 & 2.38 & 3.39 & 3.37 & 3.60 & 3.41 & 3.44 & 3.48 \\
\bottomrule
\end{tabular}
\caption{UIPersonaBench leaderboard: overall score, per-dimension scores, and per-scenario scores (1--5 scale) for all 14 models, produced by our validated Evidence-Grounded, Social-Weighted Persona Panel judge. }
\label{tab:leaderboard}
\end{table*}

\section{Experiment}

In this section, we introduce the UIPersonaBench benchmark and evaluation protocol, validate the Evidence-Grounded, Social-Weighted Persona Panel's fidelity to human judgment against alternative scoring methods, and present the resulting leaderboard over 14 state-of-the-art models.

\subsection{Experimental Setup}

\subsubsection{Benchmark.}
We construct \textbf{UIPersonaBench}, a benchmark of 500 natural-language UI-generation instructions spanning 6 common scenarios (landing/marketing, dashboard/analytics, forms/auth flow, e-commerce/content listing, social/productivity app, and mobile/widget UI), and render each instruction with every evaluated model, yielding $500 \times 14 = 7{,}000$ screenshots to be judged.

\subsubsection{Metrics.}
We evaluate a scoring method along two complementary axes against the human ground truth: score-level agreement, i.e. how closely the method's per-screenshot overall score tracks the human mean score, measured with Pearson's $r$, Spearman's $\rho$, mean absolute error (MAE), and root mean squared error (RMSE) over all $n=7{,}000$ (instruction, model) screenshots; and ranking-level agreement, i.e. whether the method recovers the same relative ordering of the 14 models as humans do, measured with Spearman's $\rho$ and Kendall's $\tau$ over the 14 per-model average scores. All 5 GenUI evaluation dimensions (Understanding, Trust\_and\_Reliance, Usability, Control, Transparency) use the same 1--5 integer scale, and a screenshot's overall score is the mean of its 5 dimension scores.

\subsubsection{Human Ground Truth.}
\label{sec:human_gt}
For every screenshot we collect 5 independent human ratings on the same 5-dimension, 1--5 scale used by our panel; a screenshot's human score is the mean over its 5 raters' per-dimension means, giving 7,000 human-anchored ground-truth scores directly comparable to our judge's output.

\subsubsection{Evaluated Models.}
We render each of the 500 instructions with 14 state-of-the-art LLMs: Claude-Opus-4.7~\cite{anthropic2026opus47}, Claude-Opus-4.6~\cite{anthropic2026opus46}, GPT-5.5~\cite{openai2026gpt55}, GPT-5.4~\cite{openai2026gpt54}, GPT-5.2~\cite{openai2025gpt52}, Gemini-2.5-Pro~\cite{comanici2025gemini}, DeepSeek-V4-Pro~\cite{xu2026deepseek}, DeepSeek-V4-Flash~\cite{xu2026deepseek}, GLM-5.1~\cite{zeng2026glm}, GLM-5~\cite{zeng2026glm}, GLM-4.7~\cite{zeng2026glm}, Kimi-K2.6~\cite{team2025kimi}, Doubao-Seed-2.0-Pro~\cite{bytedance2026seed2}, and MiniMax-M2.7~\cite{minimax2026m27}.

\subsubsection{Compared Evaluation Methods.}
We compare three evaluation methods that share the same underlying judge model, Claude-Opus-4.6, and the same screenshots, differing only in judging procedure: (i) \textbf{Naive single-pass}, a standard LLM-as-judge baseline with no persona, no panel, and a single forward pass; (ii) \textbf{Prompt-ensemble}, a stronger non-persona control in which the same screenshot is independently rated 5 times by 5 differently-framed but task-equivalent prompts (still no persona, no P-Q-A grounding, no panel), with the final score obtained by averaging the 5 per-dimension scores; since our panel also aggregates 5 independent ratings, this isolates how much of our panel's advantage is attributable to genuine persona/evidence grounding versus simply averaging multiple independent forward passes, a "prompt ensembling" effect; and (iii) \textbf{Full Evidence-Grounded, Social-Weighted Persona Panel (ESPP, ours)}, the complete three-stage pipeline, whose Stage 1 and Stage 2 are likewise instantiated with Claude-Opus-4.6, so that any fidelity difference in Table~\ref{tab:judge_fidelity} is attributable to the judging procedure and not to a change of the underlying judge.

\subsection{Main Results}

Table~\ref{tab:judge_fidelity} reports how closely each method's scores track the human ground truth at the individual-screenshot level.
Naive single-pass LLM-as-judge shows only moderate agreement with human raters, confirming that a single, persona-free judging pass systematically diverges from how a diverse population of real users perceives a GenUI screenshot. 
Simply averaging 5 differently-framed prompts without any persona or evidence grounding partially improves over the naive single pass, but recovers only about a third of the total gap to the full pipeline, indicating that little of the naive judge's disagreement with humans is attributable to noisy single-pass variance.
Most of it instead reflects the absence of diverse, grounded viewpoints. 
This shows that our full panel's advantage over a mere prompt-ensembling control comes overwhelmingly from genuine persona/evidence grounding, and only marginally from averaging multiple independent passes. 
Adding the full ESPP machinery further improves alignment on every metric, the best of all three methods. These results support our central claim: a socially-weighted panel of grounded personas is a substantially more faithful proxy for human judgment of GenUI quality than a single LLM judge, without requiring real human raters at evaluation time.

\subsection{UIPersonaBench Leaderboard}

Using our validated ESPP judge, we produce the leaderboard in Table~\ref{tab:leaderboard}. Overall scores span a narrow $3.45$--$3.73$ range, so the exact rank order among closely-clustered models is not the object of interest here and should not be over-read as a precise capability ordering; the more informative and consistent signal, detailed below, lies in the shared per-dimension and per-scenario patterns that hold across essentially every model regardless of its overall rank:

(i) \textbf{Understanding is a uniform strength, not a differentiator.} Every one of the 14 models scores highest on Understanding, nearly a point above its next-best dimension (Usability), so near-ceiling instruction comprehension is already a solved sub-problem industry-wide and contributes little to separating stronger from weaker GenUI systems.

(ii) \textbf{Transparency is a systematic, industry-wide weakness.} It is the lowest-scoring dimension for every one of the 14 models without exception, trailing Understanding by roughly two points, suggesting that explaining an interface's own state and reasoning is a shared blind spot that overall capability improvements have not yet closed.

(iii) \textbf{Dimension-level rankings reorder models relative to the overall leaderboard, unlike scenario identity.} Usability and Understanding rankings agree only moderately with the overall ranking, so a model's relative standing on these dimensions cannot be reliably inferred from its overall rank, exactly the kind of trade-off a single aggregate score would hide; per-scenario scores, by contrast, are far more consistent within each model, indicating scenario identity is a comparatively minor source of variation.

\section{Further Analysis}

The main results show that our panel tracks human judgment well, but a good aggregate score alone does not tell us whether the panel is doing so for the right reasons. This section asks five concrete questions in turn: which component of the pipeline is actually responsible for the accuracy gain; does a persona's simulated personality really change how much it revises its opinion during discussion, or is that instruction ignored; does group discussion converge to a false consensus that would defeat the purpose of simulating discussion at all; do different kinds of users actually disagree about GenUI quality in a way a single judge would erase; and can the panel be fooled by purely cosmetic interface tricks?

\subsection{Ablation Study}
\label{sec:ablation}

We ablate our Full pipeline by cumulatively removing components: Stage-2/3 (opinion dynamics and weighted aggregation), then P-Q-A grounding, then the persona panel itself (recovering the naive single-pass baseline). 
Table~\ref{tab:ablation} shows that accuracy degrades monotonically as components are removed, but not uniformly: 
removing Stage-2/3 costs little, whereas removing P-Q-A grounding alone accounts for the majority of the remaining gap to the naive baseline. 
Evidence grounding, not panel dynamics, is thus the single most load-bearing component of our pipeline, consistent with our later finding that Stage-2/3 mainly sharpens reliability and contributes only secondarily to accuracy.

\begin{table}[t]
\centering
\small
\begin{tabular}{lcccc}
\toprule
Ablation & $r$ & $\rho$ & MAE $\downarrow$ & RMSE $\downarrow$ \\
\midrule
Full (ours) & \textbf{0.922} & \textbf{0.904} & \textbf{0.110} & \textbf{0.138} \\
w/o dynamics & 0.891 & 0.857 & 0.121 & 0.152 \\
w/o grounding & 0.811 & 0.749 & 0.196 & 0.244 \\
w/o panel & 0.716 & 0.612 & 0.435 & 0.490 \\
\bottomrule
\end{tabular}
\caption{Cumulative ablation of the Full pipeline. Each row additionally removes the component named, relative to the row above.}
\label{tab:ablation}
\end{table}

\subsection{Does Trait-Derived Receptivity Drive Opinion Change?}
\label{sec:receptivity}

\begin{figure}[t]
\centering
\includegraphics[width=1\linewidth]{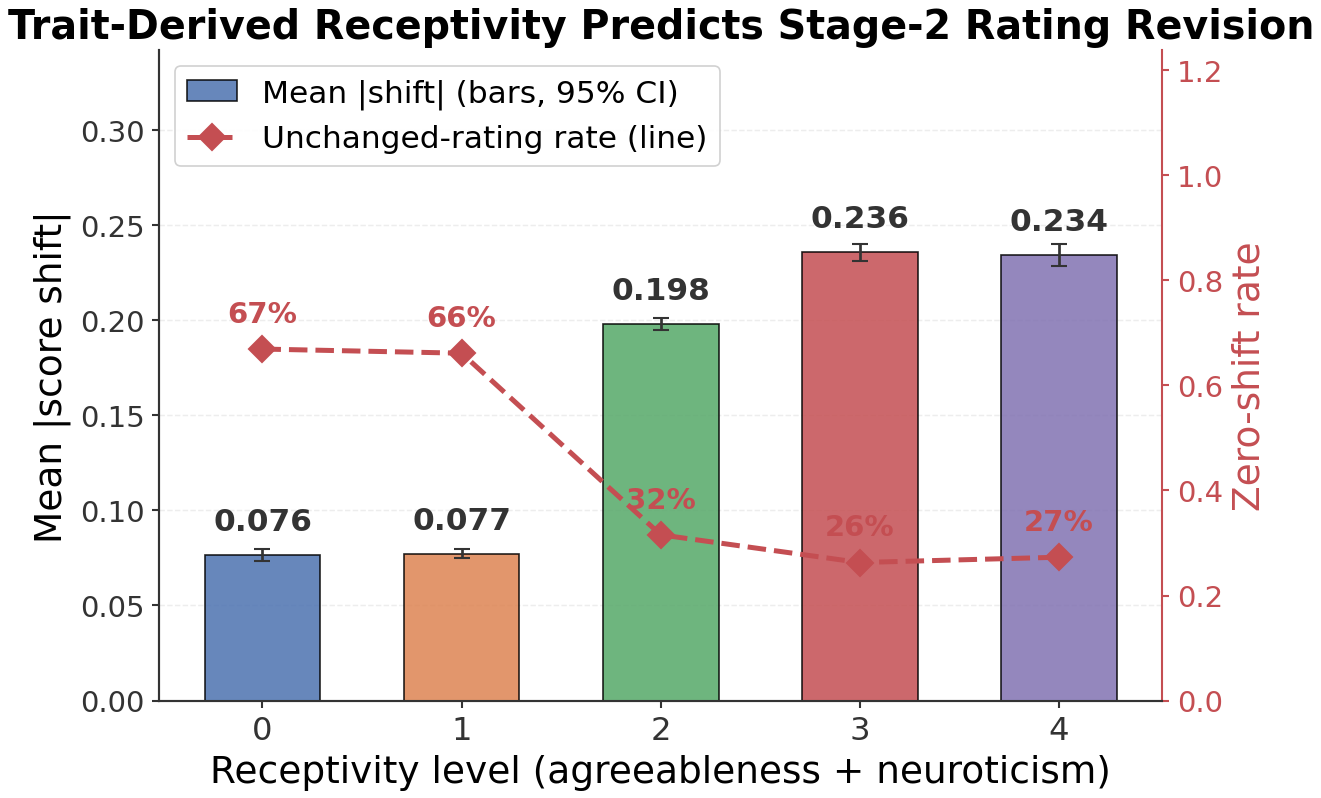}
\caption{Stage-2 rating revision as a function of a persona's trait-derived receptivity level.}
\label{fig:receptivity}
\end{figure}

Stage 2 derives each persona's susceptibility to social influence from a fixed, auditable function of Big-Five agreeableness and neuroticism, so receptivity is never left as a free hyperparameter. Whether the underlying LLM honors this instruction, as opposed to treating it as unused prompt decoration, is the empirical question we test here. If the mechanism is functioning as intended, both the magnitude of a persona's rating revision and its propensity to revise at all should increase monotonically with theorized receptivity. Figure~\ref{fig:receptivity} confirms both predictions: the mean score shift and the unchanged-rating rate move in opposite, consistently monotonic directions across all five receptivity levels, and the association is statistically robust. This indicates that the persuadability implied by a persona's personality traits goes beyond textual flavor: it measurably shapes how much a panelist's rating moves during the simulated discussion, lending behavioral validity to the opinion-dynamics mechanism.

\subsection{Opinion Dynamics: Partial, Not Total, Consensus}
\label{sec:polarization}

\begin{figure}[t]
\centering
\includegraphics[width=1\linewidth]{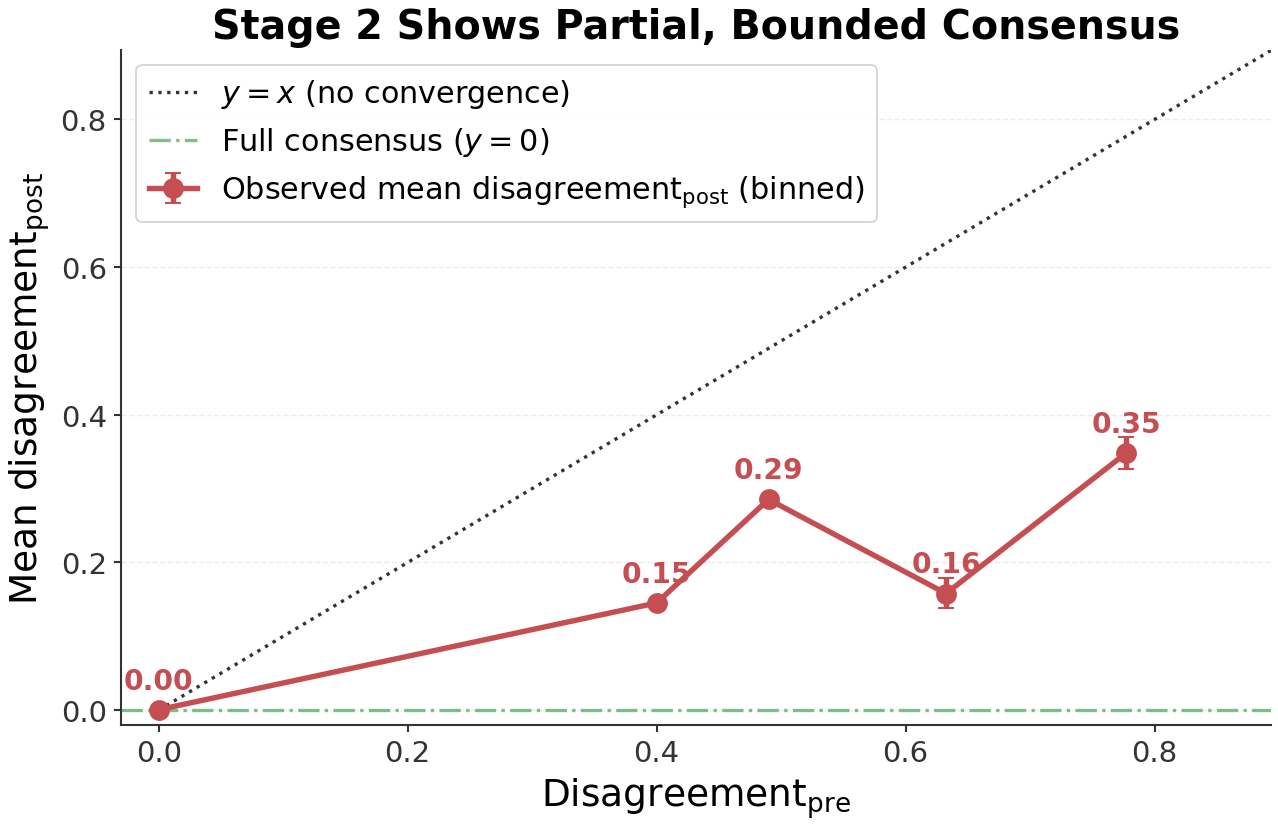}
\caption{Post-discussion vs.\ pre-discussion panel disagreement, against the $y=x$ (no convergence) and $y=0$ (full consensus) reference lines.}
\label{fig:polarization}
\end{figure}

A judge whose Stage-2 discussion always fully converges the panel would be behaviorally indistinguishable from a plain unweighted mean, undermining the motivation for simulating discussion at all. Across all $35{,}000$ (instruction, model, dimension) panels, mean within-group disagreement (standard deviation across the 5 panelists) drops from $0.283$ pre-discussion to $0.134$ post-discussion, a $52.9\%$ reduction, yet only $42.6\%$ of panels fully converge, and among the initially most divided panels (top decile of pre-discussion disagreement), $63.5\%$ still retain above-median disagreement after discussion. As Figure~\ref{fig:polarization} shows, the binned mean curve sits strictly between the two reference lines across the full range of pre-discussion disagreement, never collapsing to the $y=0$ consensus line nor sitting on the $y=x$ no-change line. This is consistent with a genuine bounded-confidence process and with neither extreme: panelists close enough in opinion and susceptible to on-topic arguments move together, while well-grounded, persona-consistent disagreement persists instead of being averaged away.

\subsection{Do Different User Subgroups See the Same Leaderboard?}
\label{sec:divergence}

\begin{figure}[t]
\centering
\includegraphics[width=1\linewidth]{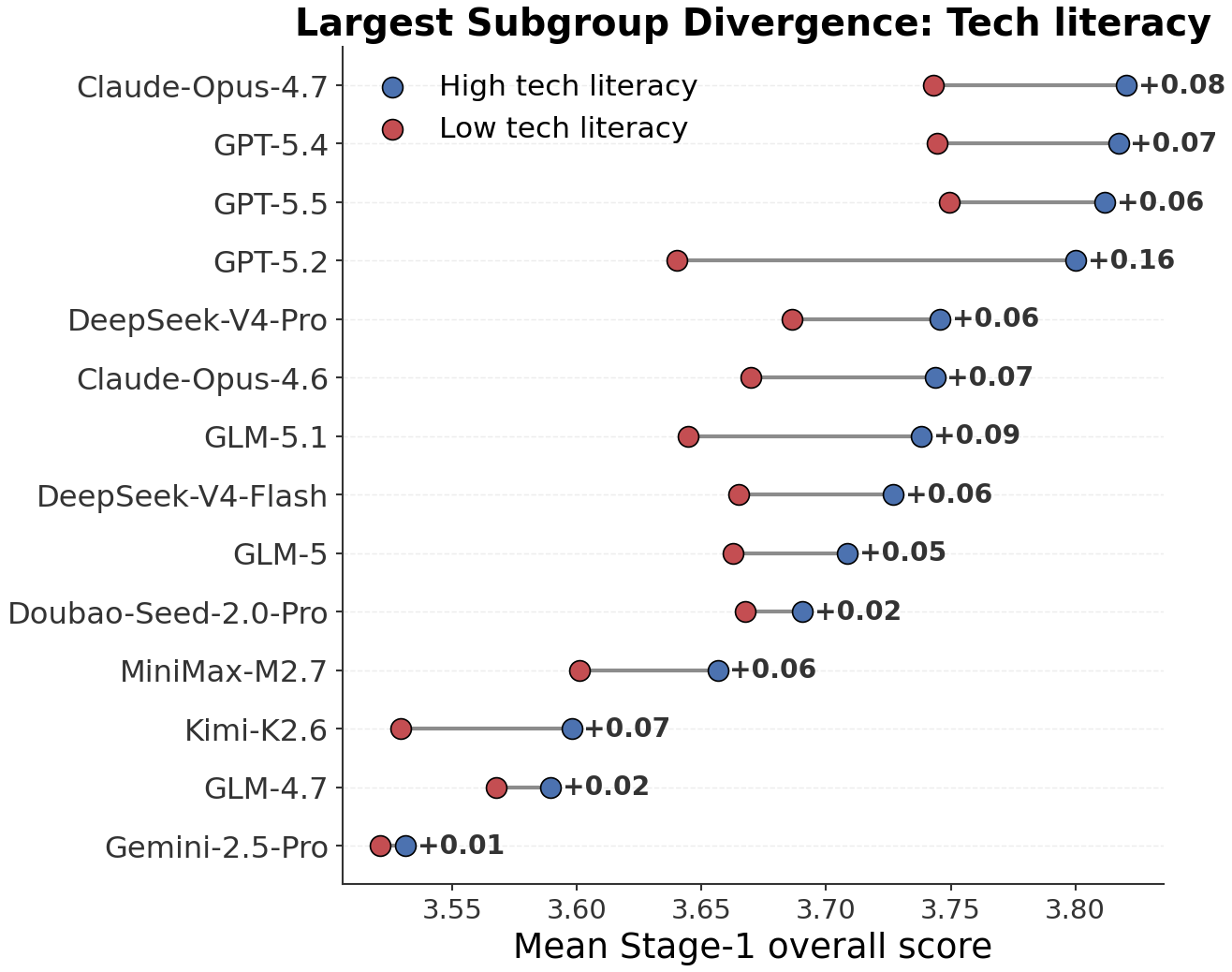}
\caption{Per-model Stage-1 score gap between subgroups, for the split with the largest leaderboard disagreement.}
\label{fig:divergence}
\end{figure}

The premise of a heterogeneous panel is that different kinds of users genuinely see GenUI quality differently. Splitting the 1,000-persona pool by tech literacy, age, domain experience, and an agreeableness-derived critical/charitable disposition, and computing each subgroup's own 14-model leaderboard from independent (pre-discussion) Stage-1 ratings, we find that overall-score rankings agree closely across subgroups, but this aggregate agreement masks substantial, localized disagreement: on the Control dimension specifically, tech-literate vs.\ non-tech-literate raters agree far less than on any other dimension, and individual models can swing by up to 6 rank positions between subgroups (Figure~\ref{fig:divergence}). To our knowledge, this is among the first results to systematically quantify \emph{dimension-localized} ranking disagreement between user subgroups in GenUI evaluation, rather than reporting only an aggregate rater-agreement figure; 
the pattern also confirms that a single homogeneous judge would obscure this disagreement, since it agrees with any one subgroup's overall ranking while erasing exactly the dimension-level split that distinguishes the subgroups in the first place.

% Further Analysis (G): systematic model-ranking disagreement across user subgroups.
% Requires: \usepackage{booktabs}
\begin{table}[t]
\centering
\small
\begin{tabular}{lccc}
\toprule
Subgroup split ($n_A$/$n_B$) & $\rho$ (overall) & $\tau$ (overall) & $\rho$ (Control) \\
\midrule
Tech literacy (257/422) & 0.846 & 0.670 & 0.70 \\
Age group (418/388) & 0.916 & 0.780 & 0.80 \\
Domain exp.\ (225/388) & 0.933 & 0.800 & 0.81 \\
Agreeableness (326/341) & 0.965 & 0.868 & 0.79 \\
\bottomrule
\end{tabular}
\caption{Rank agreement between persona subgroups' independent Stage-1 leaderboards, overall and on Control, consistently the most divergent dimension across all four splits.}
\label{tab:persona_divergence}
\end{table}
\subsection{Adversarial Robustness to Cosmetic Manipulation}
\label{sec:adversarial}

\begin{figure}[t]
\centering
\includegraphics[width=1\linewidth]{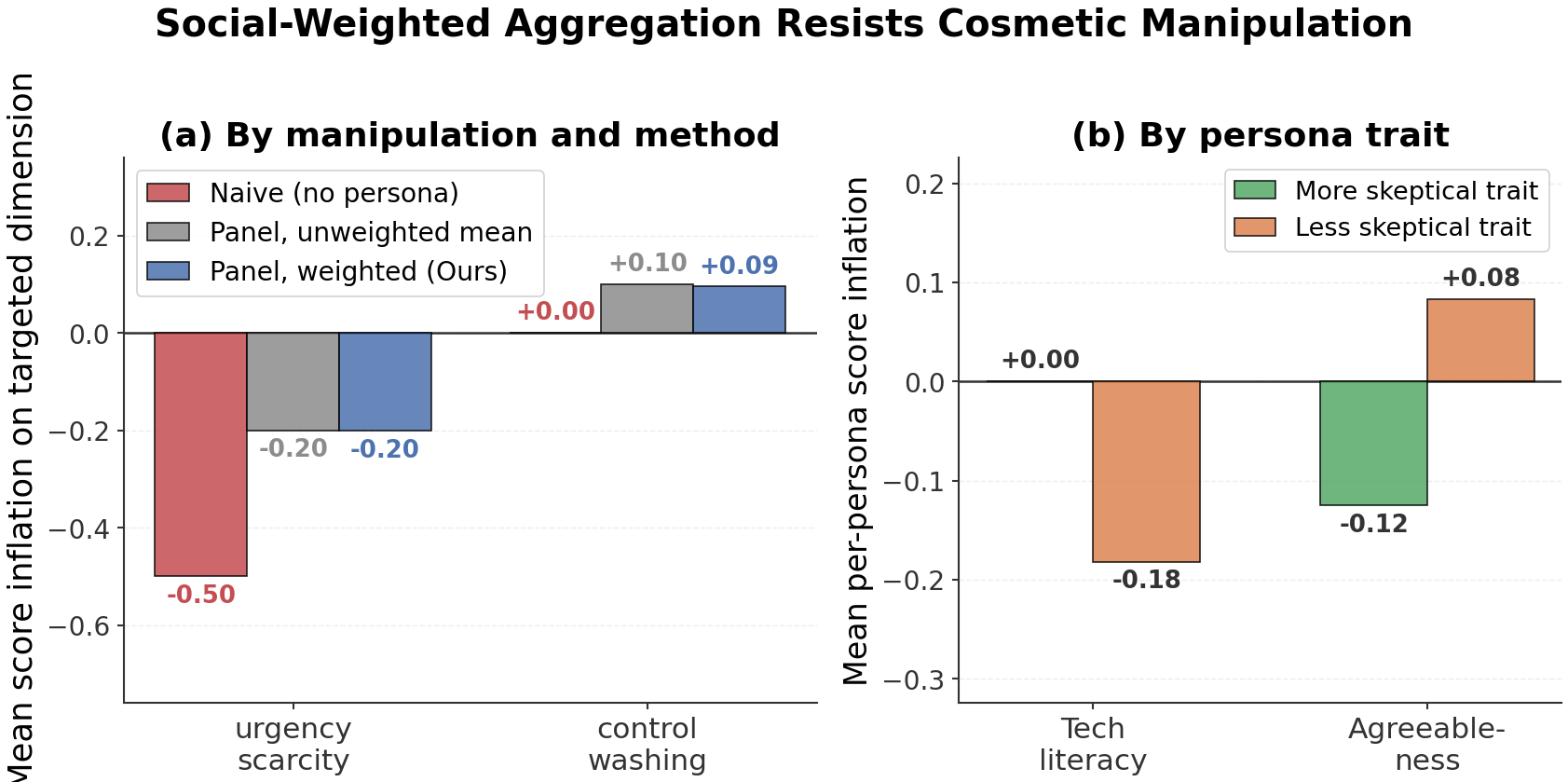}
\caption{Score change under cosmetic dark-pattern overlays. (a) By manipulation and judging method (trust badge omitted: exactly zero for all methods). (b) By persona trait, pooled across all three manipulations.}
\label{fig:adversarial}
\end{figure}

A judge that can be swayed by superficial, purely cosmetic manipulation is not measuring genuine UI quality. We pixel-edit a small set of already-judged screenshots with three fabricated dark-pattern overlays, namely a trust badge, an urgency/scarcity banner, and a control-washing banner, each targeting a specific rating dimension, and re-run real Stage-1 judging on both the original and manipulated images. On this small probe set, no method is inflated by the manipulations in a consistent, dimension-targeted way: the trust badge produces exactly zero score change for all three methods, and the remaining manipulations either move scores against the manipulator's intent or, where a small positive drift does appear, inflate panel-based scoring no more than the naive judge (Figure~\ref{fig:adversarial}a). Pooling per-persona score changes across all three manipulations, any residual susceptibility concentrates in traits associated with lower skepticism: low tech-literacy and high-agreeableness personas both show larger inflation than their more critical counterparts (Figure~\ref{fig:adversarial}b), consistent with a socially-weighted panel being tunable toward more critical raters for added robustness. 

\section{Conclusion}
\label{sec:conclusion}

A single LLM judge is not merely inaccurate: it is structurally incapable of representing disagreement, since it collapses every rater's viewpoint into one score before that disagreement can ever be observed. 
We presented ESPP, a three-stage method that grounds diverse personas in prior evidence, lets them deliberate, and aggregates their judgments while retaining each panelist's individual rating. On UIPersonaBench, ESPP tracks human judgment far more closely than a naive single-pass judge, driven mainly by persona and evidence grounding rather than sampling variance, and its retained individual ratings further reveal real, dimension-localized disagreement across user subgroups that a single-score judge would erase. We see this as a step toward automatic judges that keep sources of disagreement legible rather than silently collapsing them away into a single opaque number.

\bibliography{aaai2027}


\end{document}